\documentclass[10pt,twocolumn,letterpaper]{article}

\usepackage{wacv}
\usepackage{times}
\usepackage{epsfig}
\usepackage{graphicx}
\usepackage{amsmath}
\usepackage{amssymb}
\usepackage{booktabs}
\usepackage{bm}

\def\wacvPaperID{215}
\wacvfinalcopy
\ifwacvfinal
    \def\assignedStartPage{9876}
\fi

\ifwacvfinal
    \usepackage[breaklinks=true,bookmarks=false]{hyperref}
\else
    \usepackage[pagebackref=true,breaklinks=true,colorlinks,bookmarks=false]{hyperref}
\fi
\ifwacvfinal
\else
\fi

\begin{document}

\title{Efficient Attention: Attention with Linear Complexities}

\author{
    Shen Zhuoran$^\dagger$\thanks{Work during internship at SenseTime.} \\
    Independent Researcher \\
    4244 University Way NE \#85406, Seattle, WA 98105, United States \\
    {\tt\small cmsflash99@gmail.com}
    \and
    Zhang Mingyuan\thanks{Equal contribution.}, Zhao Haiyu, Yi Shuai \\
    SenseTime International \\
    182 Cecil Street, \#36-02 Frasers Tower, Singapore 069547 \\
    {\tt\small zhangmingyuan,zhaohaiyu,yishuai@sensetime.com}
    \and
    Li Hongsheng \\
    The Chinese University of Hong Kong \\
    Sha Tin, Hong Kong \\
    {\tt\small hsli@ee.cuhk.edu.hk}
}

\maketitle

\begin{abstract}
    Dot-product attention has wide applications in computer vision and natural language processing. However, its memory and computational costs grow quadratically with the input size. Such growth prohibits its application on high-resolution inputs. To remedy this drawback, this paper proposes a novel efficient attention mechanism equivalent to dot-product attention but with substantially less memory and computational costs. Its resource efficiency allows more widespread and flexible integration of attention modules into a network, which leads to better accuracies. Empirical evaluations demonstrated the effectiveness of its advantages. Efficient attention modules brought significant performance boosts to object detectors and instance segmenters on MS-COCO 2017. Further, the resource efficiency democratizes attention to complex models, where high costs prohibit the use of dot-product attention. As an exemplar, a model with efficient attention achieved state-of-the-art accuracies for stereo depth estimation on the Scene Flow dataset. Code is available at \url{https://github.com/cmsflash/efficient-attention}.
\end{abstract}
\section{Introduction}
\label{sec:intro}

Dot-product attention \cite{attention,transformer,nonlocal} is a prevalent mechanism in neural networks for long-range dependency modeling, a key challenge to deep learning that convolution and recurrence struggle to solve. The mechanism computes the response at every position as a weighted sum of features at all positions in the previous layer. In contrast to the limited receptive fields of convolution or the recurrent layer, dot-product attention expands the receptive field to the entire input in one pass. Using dot-product attention to efficiently model long-range dependencies allows convolution and recurrence to focus on local dependency modeling, in which they specialize. The non-local module \cite{nonlocal}, an adaptation of dot-product attention for computer vision, achieved state-of-the-art performance on video classification \cite{nonlocal} and generative adversarial image modeling \cite{sagan,biggan} and demonstrated significant improvements on object detection \cite{nonlocal}, instance segmentation \cite{nonlocal}, person re-identification \cite{video-reid}, image de-raining \cite{deraining}, \etc.

However, global dependency modeling on large inputs (\eg long sequences, high-resolution images, large videos) remains an open problem. The quadratic\footnote{The complexities are quadratic with respect to the spatiotemporal size of the input, which is quartically \wrt the side length of a 2D feature map, or sextically \wrt the dimension of a 3D feature volume.} memory and computational complexities with respect to the input size of dot-product attention inhibits its application on large inputs. For instance, a non-local module uses over 1 GB of GPU memory and over 25 GMACC\footnote{MACC stands for multiply-accumulation. 1 MACC means 1 multiplication and 1 addition operation.} of computation for a 64-channel $128 \times 128$ feature map or over 68 GB and over 1.6 TMACC for a 64-channel $64 \times 64 \times 32$ 3D feature volume (\eg for depth estimation or video tasks). The high memory and computational costs constrain the application of dot-product attention to the low-resolution parts of models \cite{nonlocal,sagan,biggan} and prohibits its use for resolution-sensitive or resource-hungry tasks.

\begin{figure*}[t]
    \centering
    \includegraphics[width=\linewidth]{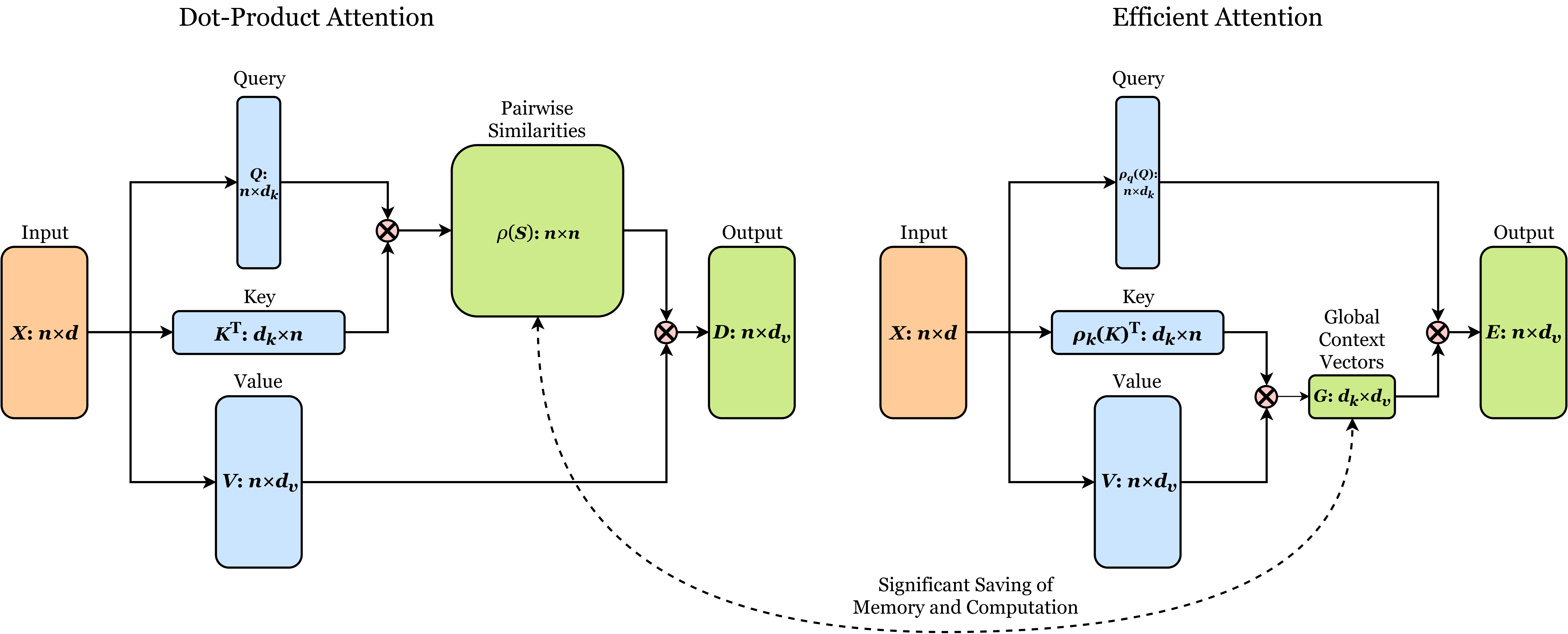}
    \caption{\textbf{Illustration of the architecture of dot-product and efficient attention.} Each box represents an input, output, or intermediate matrix. Above it is the name of the corresponding matrix, and inside are the variable name and the size of the matrix. $\rho, \rho_q, \rho_k$ are the normalizers on $\bm{S}, \bm{Q}, \bm{K}$, respectively. $n, d, d_k, d_v$ are the input size and the dimensionalities of the input, the keys, and the values, respectively. $\bigotimes$ denotes matrix multiplication. When $\rho, \rho_q, \rho_k$ implement scaling normalization, the efficient attention mechanism is mathematically equivalent to dot-product attention. When they implement softmax normalization, the two mechanisms are approximately equivalent.}
    \label{fig:ca-ea}
\end{figure*}

The need for global dependency modeling on large inputs motivates the exploration for a resource-efficient attention mechanism. An investigation into the non-local module revealed an intriguing discovery. As Figure \ref{fig:ca-ea} shows, putting aside the normalization, dot-product attention involves two consecutive matrix multiplications. The first one ($\bm{S} = \bm{Q}\bm{K}^\mathsf{T}$) computes pairwise similarities between pixels and forms per-pixel attention maps. The second ($\bm{D} = \bm{S}\bm{V}$) aggregates the values $\bm{V}$ by the per-pixel attention maps to produce the output. Since matrix multiplication is associative, switching the order from $(\bm{Q}\bm{K}^\mathsf{T})\bm{V}$ to $\bm{Q}(\bm{K}^\mathsf{T}\bm{V})$ has no impact on the effect but changes the complexities from $O(n^2)$ to $O(d_kd_v)$, for $n$ the input size and $d_k, d_v$ the dimensionalities of the keys and the values, respectively. This change removes the $O(n^2)$ terms in the complexities of the module, making it linear in complexities. Further, $d_kd_v$ is significantly less than $n^2$ in practical cases, hence this new term will not become a new bottleneck. Therefore, switching the order of multiplication to $\bm{Q}(\bm{K}^\mathsf{T}\bm{V})$ results in a substantially more efficient mechanism, which this paper names \textit{efficient attention}.

The new mechanism is mathematically equivalent to dot-product attention with scaling normalization and approximately equivalent with softmax normalization. Experiments empirically verified that when the equivalence is approximate, it does not impact accuracies. In addition, experiments showed that its efficiency allows the integration of more attention modules into a network and integration into high-resolution parts of a network, which lead to significantly higher accuracies. Further, experiments demonstrated that efficient attention democratizes attention to tasks where dot-product attention is inapplicable due to resource constraints.

Another discovery is that efficient attention brings a new interpretation to the attention mechanism. Assuming the keys are of dimensionality $d_k$ and the input size is $n$, one can interpret the $d_k \times n$ key matrix as $d_k$ template attention maps, each corresponding to a semantic aspect of the input. Then, the query at each pixel is $d_k$ coefficients for each of the $d_k$ template attention maps, respectively. Under this interpretation, efficient and dot-product attention differs in that dot-product attention first synthesizes the pixel-wise attention maps from the coefficients and lets each pixel aggregate the values with its own attention map, while efficient attention first aggregates the values by the template attention maps to form template outputs (\ie global context vectors) and lets each pixel aggregate the template outputs.

The principal contribution of this paper is the efficient attention mechanism, which:
\begin{enumerate}
    \item has linear memory and computational complexities with respect to the size of the input;
    \item possesses the same representational power as the prevalent dot-product attention mechanism;
    \item allows the integration of more attention modules into a neural network and into higher-resolution parts of the network, which brings substantial performance boosts to tasks such as object detection and instance segmentation (on MS-COCO 2017); and
    \item facilitates the application of attention on resource-hungry tasks, such as stereo depth estimation (on the Scene Flow dataset).
\end{enumerate}

\section{Related works}

\subsection{Dot-product attention}

\cite{attention} proposed the initial formulation of the dot-product attention mechanism to improve word alignment in machine translation. Successively, \cite{transformer} proposed to completely replace recurrence with attention and named the resultant architecture the Transformer. The Transformer architecture is highly successful on sequence tasks. They hold the state-of-the-art records on virtually all tasks in natural language processing \cite{bert,gpt2,xlnet} and is highly competitive on end-to-end speech recognition \cite{speech-tfm,deep-san}. \cite{nonlocal} first adapted dot-product attention for computer vision and proposed the non-local module. They achieved state-of-the-art performance on video classification and demonstrated significant improvements on object detection, instance segmentation, and pose estimation. Subsequent works applied it to various fields in computer vision, including image restoration \cite{nlrn}, video person re-identification \cite{video-reid}, generative adversarial image modeling \cite{sagan,biggan}, image de-raining \cite{deraining}, and few-shot learning \cite{can,silco}, \etc.

Efficient attention mainly builds upon the version of dot-product attention in the non-local module. Following \cite{nonlocal}, the team conducted most experiments on object detection and instance segmentation. The paper compares the resource efficiency of the efficient attention module against the non-local module under the same performance and their performance under the same resource constraints.

\subsection{Scaling attention}

Besides dot-product attention, there are a separate set of techniques the literature refers to as attention. This section refers to them as scaling attention. While dot-product attention is effective for global dependency modeling, scaling attention focuses on emphasizing important features and suppressing uninformative ones. For example, the squeeze-and-excitation (SE) module \cite{senet} uses global average pooling and a linear layer to compute a scaling factor for each channel and then scales the channels accordingly. SE-enhanced models achieved state-of-the-art performance on image classification and substantial improvements on scene segmentation and object detection. On top of SE, CBAM \cite{cbam} added global max pooling beside global average pooling and an extra spatial attention submodule. GCNet \cite{gcnet} proposes to replace the global average pooling by an adaptive pooling layer, which uses a linear layer to compute the weight for each position. These follow-up methods further improves upon the performance of SE \cite{senet}.

Despite both names containing attention, dot-product attention and scaling attention are two separate sets of techniques with highly divergent goals. When appropriate, one might take both techniques and let them work in conjunction. Therefore, it is unnecessary to make comparison of efficient attention with scaling attention techniques.

\subsection{Efficient non-local operations}

Recent literature proposed several methods for efficient non-local operations. LatentGNN \cite{latentgnn} proposes to approximate the single $n \times n$ affinity matrix in the non-local \cite{nonlocal} module by the product of three lower-rank matrices. In comparison, efficient attention is not an approximation of the non-local module, but is mathematically equivalent (using scaling normalization). In addition, there is a one-to-one mapping between the structural components of the non-local module and the efficient attention module. Therefore, in any field where the non-local module succeeded, one can guarantee the applicability of efficient attention as a drop-in replacement with substantially improved performance-cost trade-off.

CGNL \cite{cgnl} proposes to flatten the height, width, and channel dimensions to a $hwc$-dimensional vector, applies a kernel function to expand the dimensionality to $hwc \times (p + 1)$, for $p$ the degree of Taylor expansion, and models global dependencies in that space. However, after flattening the input into a vector, the feature at each position becomes a scalar, which encodes limited information for interaction modeling. In contrast, efficient attention preserves a vector representation at each pixel and is capable to model richer interactions.

Section \ref{sec:vs-comp} presents empirical comparison between efficient attention and these competing methods in detail, which shows that efficient attention outperforms each of them.

\section{Method}
\label{sec:method}

\subsection{A revisit of dot-product attention}
\label{sec:ca}

Dot-product attention is a mechanism for long-range interaction modeling in neural networks. For each input feature vector $\bm{x}_i \in \mathbb{R}^d$ that corresponds to the $i$-th position, dot-product attention first uses three linear layers to convert $\bm{x}_i$ into three feature vectors, \ie, the query $\bm{q}_i \in \mathbb{R}^{d_k}$, the key $\bm{k}_i \in \mathbb{R}^{d_k}$, and the value $\bm{v}_i \in \mathbb{R}^{d_v}$. The queries and keys must have the same feature dimension $d_k$. One can measure the similarity between the $i$-th query and the $j$-th key as $\rho(\bm{q}_i^\mathsf{T}\bm{k}_j)$, where $\rho$ is a normalization function. In general, the similarities are asymmetric, since the queries and keys are the outputs of two separate layers. The dot-product attention module calculates the similarities between all pairs of positions. Using the similarities
as weights, position $i$ aggregates the values from all positions via weighted summation to obtain its output feature.

If one represents all $n$ positions' queries, keys, and values in matrix forms as $\bm{Q} \in \mathbb{R}^{n \times d_k}$, $\bm{K} \in \mathbb{R}^{n \times d_k}, \bm{V} \in \mathbb{R}^{n\times d_v}$, respectively, the output of dot-product attention is
\begin{equation}
\label{eq:ca}
	\bm{D}(\bm{Q}, \bm{K}, \bm{V}) = \rho\left(\bm{Q}\bm{K}^\mathsf{T}\right)\bm{V}.
\end{equation}
The normalization function has two common choices:
\begin{align}
\begin{split}
\label{eq:ca-norm}
    \text{Scaling: } & \rho(\bm{Y}) = \frac{\bm{Y}}{n}, \\
    \text{Softmax: } & \rho(\bm{Y}) = \sigma_\text{row}(\bm{Y}),
\end{split}
\end{align}
where $\sigma_\text{row}$ denotes applying the softmax function along each row of matrix $\bm{Y}$.
An illustration of the dot-product attention module is in Figure \ref{fig:ca-ea} (left).

The critical drawback of this mechanism is its resource demands. Since it computes a similarity between each pair of positions, there are $n^2$ such similarities, which results in $O(n^2)$ memory complexity and $O(d_kn^2)$ computational complexity. Therefore, dot-product attention's resource demands get prohibitively high on large inputs. In practice, application of the mechanism is only possible on low-resolution features.

\subsection{Efficient attention}
\label{sec:ea}

Observing the critical drawback of dot-product attention, this paper proposes the efficient attention mechanism, which is mathematically equivalent to dot-product attention but substantially faster and more memory efficient. In efficient attention, the individual feature vectors $\bm{X} \in \mathbb{R}^{n \times d}$ still pass through three linear layers to form the queries $\bm{Q} \in \mathbb{R}^{n \times d_k}$, keys $\bm{K} \in \mathbb{R}^{n \times d_k}$, and values $\bm{V} \in \mathbb{R}^{n \times d_v}$. However, instead of interpreting the keys as $n$ feature vectors in $\mathbb{R}^{d_k}$, the module regards them as $d_k$ single-channel feature maps. Efficient attention uses each of these feature maps as a weighting over all positions and aggregates the value features through weighted summation to form a global context vector. The name reflects the fact that the vector does not correspond to a specific position, but is a global description of the input features.

The following equation characterizes the efficient attention mechanism:
\begin{equation}
\label{eq:ea}
	\bm{E}(\bm{Q}, \bm{K}, \bm{V}) = \rho_q(\bm{Q})\left(\rho_k(\bm{K})^\mathsf{T}\bm{V}\right),
\end{equation}
where $\rho_q$ and $\rho_k$ are normalization functions for the query and key features, respectively. The implementation of the same two normalization methods as for dot-production attention are
\begin{align}
\begin{split}
\label{eq:ea-norm}
    \text{Scaling: } & \rho_q(\bm{Y}) = \rho_k(\bm{Y}) = \frac{\bm{Y}}{\sqrt{n}}, \\
    \text{Softmax: } & \rho_q(\bm{Y}) = \sigma_\text{row}(\bm{Y}), \\
    & \rho_k(\bm{Y}) = \sigma_\text{col}(\bm{Y}),
\end{split}
\end{align}
where $\sigma_\text{row}, \sigma_\text{col}$ denote applying the softmax function along each row or column of matrix $\bm{Y}$, respectively.

The efficient attention module is a concrete implementation of the mechanism for computer vision data. For an input feature map $\bm{\mathsf{X}} \in \mathbb{R}^{h \times w \times d}$, the module flattens it to a matrix $\bm{X} \in \mathbb{R}^{hw \times d}$, applies the efficient attention mechanism on it, and reshapes the result to $h \times w \times d_v$. If $d_v \ne d$, it further applies a 1x1 convolution to restore the dimensionality to $d$. Finally, it adds the resultant features to the input features to form a residual structure.

\subsection{Equivalence between dot-product and efficient attention}
\label{sec:proof}

Following is a formal proof of the equivalence between dot-product and efficient attention when using scaling normalization. Substituting the scaling normalization formula in Equation \eqref{eq:ca-norm} into Equation \eqref{eq:ca} gives
\begin{equation}
\label{eq:ca-scaling}
    \bm{D}(\bm{Q}, \bm{K}, \bm{V}) = \frac{\bm{Q}\bm{K}^\mathsf{T}}{n}\bm{V}.
\end{equation}
Similarly, plugging the scaling normalization formulae in Equation \eqref{eq:ea-norm} into Equation \eqref{eq:ea} results in
\begin{equation}
\label{eq:ea-scaling}
    \bm{E}(\bm{Q}, \bm{K}, \bm{V}) = \frac{\bm{Q}}{\sqrt{n}}\left(\frac{\bm{K}^\mathsf{T}}{\sqrt{n}}\bm{V}\right).
\end{equation}

Since scalar multiplication is commutative with matrix multiplication and matrix multiplication is associative, we have
\begin{align}
\begin{split}
\label{eq:proof}
    \bm{E}(\bm{Q}, \bm{K}, \bm{V}) &= \frac{\bm{Q}}{\sqrt{n}}\left(\frac{\bm{K}^\mathsf{T}}{\sqrt{n}}\bm{V}\right) \\
    &= \frac{1}{n}\bm{Q}\left(\bm{K}^\mathsf{T}\bm{V}\right) \\
    &= \frac{1}{n}\left(\bm{Q}\bm{K}^\mathsf{T}\right)\bm{V} \\
    &= \frac{\bm{Q}\bm{K}^\mathsf{T}}{n}\bm{V}.
\end{split}
\end{align}
Comparing Equations \eqref{eq:ca-scaling} and \eqref{eq:proof}, we get
\begin{equation}
\label{eq:qed}
    \bm{E}(\bm{Q}, \bm{K}, \bm{V}) = \bm{D}(\bm{Q}, \bm{K}, \bm{V}).
\end{equation}
Thus, the proof is complete.

The above proof works for the softmax normalization variant with one caveat. The two softmax operations on $\bm{Q}, \bm{K}$ are not exactly equivalent to the single softmax on $\bm{Q}\bm{K}^\mathsf{T}$. However, they closely approximate the effect of the original softmax function. The critical property of $\sigma_\text{row}\left(\bm{Q}\bm{K}^\mathsf{T}\right)$ is that each row of it sums up to $1$ and represents a normalized attention distribution over all positions. The matrix $\sigma_\text{row}(\bm{Q})\sigma_\text{col}
(\bm{K})^\mathsf{T}$ shares this property. Therefore, the softmax variant of efficient attention is a close approximation of that variant of dot-product attention. Section \ref{sec:vs} demonstrates this claim empirically.

\subsection{Interpretation of efficient attention}
\label{sec:interpretation}

Efficient attention brings a new interpretation of the attention mechanism. In dot-product attention, selecting position $i$ as the reference position, one can collect the similarities of all positions to position $i$ and form an attention map $\bm{s}_i$ for that position. The attention map $\bm{s}_i$ represents the degree to which position $i$ attends to each position $j$ in the input. A higher value for position $j$ on $\bm{s}_i$ means position $i$ attends more to position $j$. In dot-product attention, every position $i$ has such an attention map $\bm{s}_i$, which the mechanism uses to aggregate the values $\bm{V}$ to produce the output at position $i$.

In contrast, efficient attention does not generate an attention map for each position. Instead, it interprets the keys $\bm{K} \in \mathbb{R}^{n \times d_k}$ as $d_k$ attention maps $\bm{k}^\mathsf{T}_j$. Each $\bm{k}^\mathsf{T}_j$ is a global attention map that does not correspond to any specific position. Instead, each of them corresponds to a semantic aspect of the entire input. For example, one such attention map might cover the persons in the input. Another might correspond to the background. Section \ref{sec:vis} gives several concrete examples. Efficient attention uses each $\bm{k}_j^\mathsf{T}$ to aggregate the values $\bm{V}$ and produce a global context vector $\bm{g}_j$. Since $\bm{k}^\mathsf{T}_j$ describes a global, semantic aspect of the input, $\bm{g}_j$ also summarizes a global, semantic aspect of the input. Then, position $i$ uses $\bm{q}_i$ as a set of coefficients over $\bm{g}_0, \bm{g}_1, \ldots, \bm{g}_{d_k - 1}$. Using the previous example, a person pixel might place a large weight on the global context vector for persons to refine its representation. A pixel at the boundary of an object might have large weights on the global context vectors for both the object and the background to enhance the contrast.

\subsection{Efficiency advantage}
\label{sec:efficiency}

This section analyzes the efficiency advantage of efficient attention over dot-product attention in memory and computation. The reason behind the efficiency advantage is that efficient attention does not compute a similarity between each pair of positions, which would occupy $O(n^2)$ memory and require $O(d_kn^2)$ computation to generate. Instead, it only generates $d_k$ global context vectors in $\mathbb{R}^{d_v}$. This change eliminates the $O(n^2)$ terms from both the memory and computational complexities of the module. Consequently, efficient attention has $O(dn + d^2)$ memory and $O(d^2n)$ computational complexities, assuming the common setting of $d_v = d, d_k = \frac{d}{2}$. Table \ref{tab:complexity} shows complexity formulae of the efficient attention module and the non-local module (using dot-product attention) in detail. In computer vision, this complexity difference is substantial. Firstly, the input size $n$ is quadratic in image side length and often very large in practice. Secondly, $d_k$ is a parameter of the module, which the designer of a network can tune to meet different resource requirements. Section \ref{sec:channels} shows that, within a reasonable range, this parameter has minimal impact on performance. This result means that an efficient attention module can typically have a small $d_k$, which further increases its efficiency advantage over dot-product attention.

\begin{table*}
    \centering    
    \caption{\textbf{Comparison of resource usage of the efficient attention and non-local modules.} This table assumes that $d_v = d, d_k = \frac{d}{2}$, which is a common setting in the literature for dot-product attention}
    \label{tab:complexity}
    \vspace{5pt}
    \begin{tabular}{l@{\hskip 12pt}r@{\hskip 12pt}r}
        \toprule
        Metric & Efficient attention module & Non-local module \\
        \midrule
        Memory (floats) & $4dn + \frac{d^2}{2}$ & $4dn + n^2$ \\
        Computation (MACC) & $(6d^2 + d)n$ & $(4d^2 + d)n + 3dn^2$ \\
        \addlinespace
        Memory complexity & $O(dn + d^2)$ & $O(dn + n^2)$ \\
        Comp. complexity & $O(d^2n)$ & $O(d^2n + dn^2)$ \\
        \bottomrule
    \end{tabular}
\end{table*}

\begin{figure}[t]
    \centering
    \includegraphics[width=\linewidth]{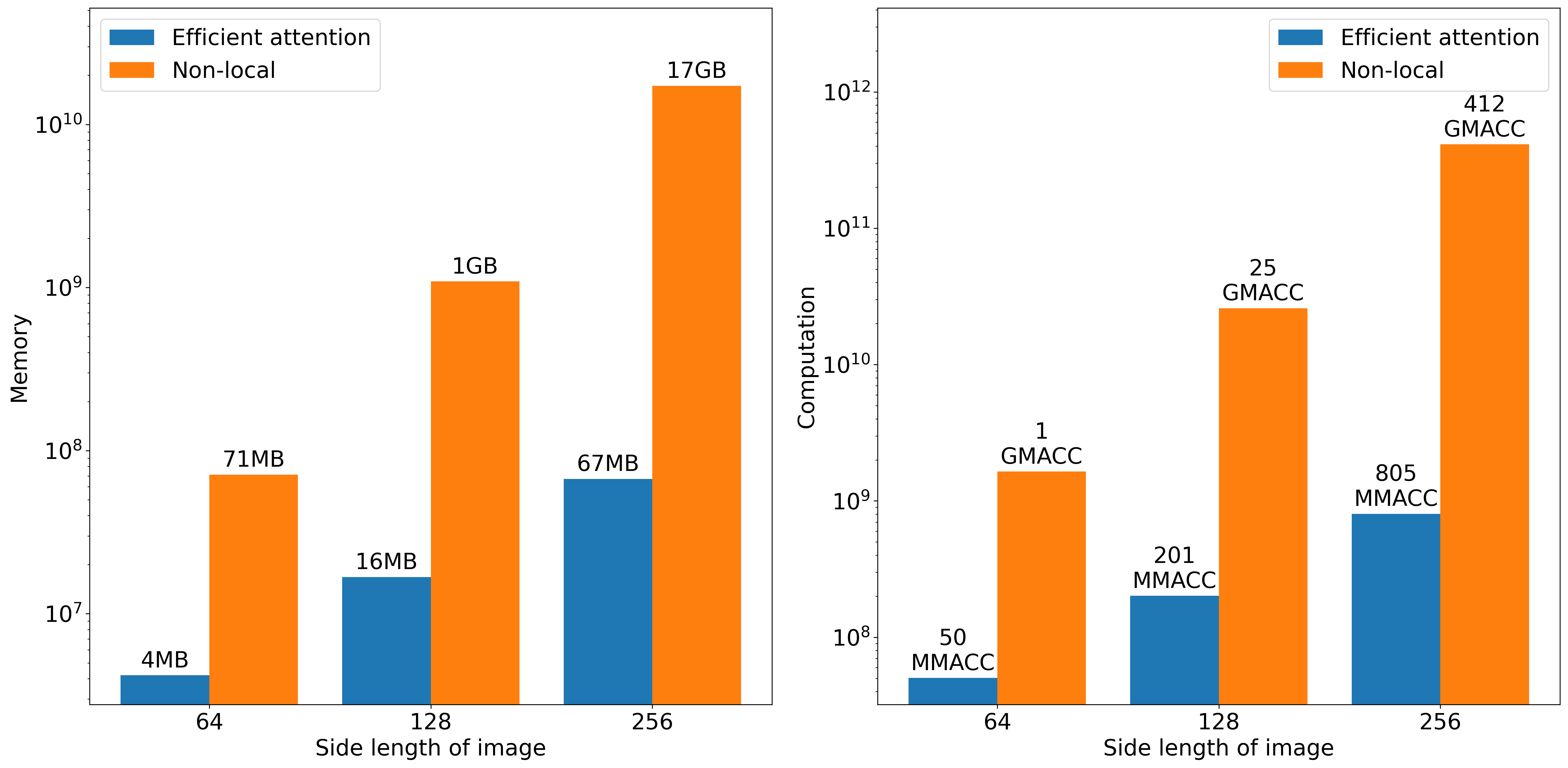}
    \caption{\textbf{Resource requirements under different input sizes.} The blue and orange bars depict the resource requirements of the efficient attention and non-local modules, respectively. The calculation assumes $d = d_v = 2d_k = 64$. This is a typical setting of self-attention for computer vision. The figure is in log scale.}
    \label{fig:complexity}
\end{figure}

The rest of this section will give several concrete examples comparing the resource demands of the efficient attention and non-local modules. Figure \ref{fig:complexity} compares their resource consumption for image features with different sizes. Directly substituting the non-local module on the $64 \times 64$ feature map in SAGAN \cite{sagan} yields a 17-time saving of memory and 33-time saving of computation. The gap widens rapidly with the increase of the input size. For a $256 \times 256$ feature map, a non-local module would require impractical amounts of memory (17.2 GB) and computation (413 GMACC). With the same input size, an efficient attention module uses 1/257 the memory and 1/513 the computation. The difference is more prominent for 3D features. For a tiny $28 \times 28 \times 4$ feature volume, an efficient attention module uses less than 1/10 the memory and computation in comparison to a non-local module. On a larger $64 \times 64 \times 32$ feature volume, a non-local module requires 513 times the memory and 1025 times the computation of an efficient attention module.

\section{Experiments on the MS-COCO task suite}
\label{sec:coco}

This section presents comparison experiments on the MS-COCO 2017 dataset for object detection and instance segmentation. The baseline is a ResNet-50 Mask R-CNN with a 5-level feature pyramid \cite{fpn}. More architectural details are in Appendix \ref{sec:arch}. The backbones initialize from ImageNet pretrainings. All other modules use random initialization. All models trained for 24 epochs on 32 NVIDIA TITAN Xp GPUs. The batch size is 64. The learning rate is $1.25 \times 10^{-4}$ at the beginning of training and drops by a factor of 10 at the start of the 18th and 21st epochs. The experiments by default use softmax normalization, $d_k = d_v = 64$, and reprojection to the original number of channels.

\subsection{Comparison experiments}
\label{sec:vs}

\subsubsection{Comparison with the non-local module}

Table \ref{tab:vs-ca} reports the comparison against the non-local module. Efficient attention achieves substantially better performance-cost trade-off. As rows \textit{res3} to \textit{fpn5} show, inserting an efficient attention module or a non-local module at the same location in a network has nearly identical effects on the performance, while efficient attention uses orders of magnitude less resources. Rows \textit{res3-4+fpn3-5} and \textit{res3-4+fpn1-5} show that under the same resource cap (TITAN Xp GPU, 12 GB VRAM), efficient attention achieves significantly better performance. Note that \textit{res3-4+fpn3-5} is the best configuration that fits in memory for non-local modules. Further inserting non-local modules to \textit{fpn1} or \textit{fpn2} would require gigabytes of memory \textit{per example}.

\begin{table*}
    \centering
    \caption{\textbf{Comparison between the efficient attention and non-local modules on MS-COCO 2017 object detection and instance segmentation.} \textit{Box}, \textit{mask}, \textit{mem.}, and \textit{comp.} stand for box AP, mask AP, memory (in bytes), and computation (in MACC), respectively. \textit{Mem.} and \textit{comp.} only count the attention module(s). \textit{res\{x\}} and \textit{fpn\{x\}} indicate inserting attention modules after the $x$-th ResBlock group or FPN level $x$, respectively. \textit{res\{x-y\}} and \textit{fpn\{x-y\}} similarly mean inserting after every ResBlock group or FPN level within the range $[x, y]$}
    \label{tab:vs-ca}
    \vspace{5pt}
    \begin{tabular}{lrrrrrrrrr}
        \toprule
         & \multicolumn{4}{c}{EA module} & \multicolumn{4}{c}{Non-local module} & \\
        \cmidrule(lr){2-5}\cmidrule(lr){6-9}
        Layer(s) & Box & Mask & Mem. & Comp. & Box & Mask & Mem. & Comp. & Input size \\
        \midrule
        None & 39.4 & 35.1 & 0 & 0 & 39.4 & 35.1 & 0 & 0 & N/A \\
        \addlinespace
        res3 & 40.2 & 36.0 & 41.3 M & 1.21 G & 40.3 & 35.9 & 122 M & 3.74 G & $56 \times 80$ \\
        res4 & 40.2 & 35.9 & 19.5 M & 596 M & 40.1 & 36.0 & 24.5 M & 748 M & $28 \times 40$ \\
        fpn1 & 39.9 & 35.8 & 220 M & 5.28 G & OOM & OOM & 20.8 G & 662 G & $224 \times 320$ \\
        fpn2 & 39.7 & 35.7 & 55.1 M & 1.32 G & OOM & OOM & 1.34 G & 42.3 G & $112 \times 160$ \\
        fpn3 & 39.7 & 35.5 & 13.8 M & 330 M & 39.8 & 35.5 & 94.0 M & 2.86 G & $56 \times 80$ \\
        fpn4 & 39.7 & 35.4 & 3.46 M & 82.6 M & 39.5 & 35.3 & 8.46 M & 234 M & $28 \times 40$ \\
        fpn5 & 39.6 & 35.3 & 877 K & 20.6 M & 39.4 & 35.2 & 1.17 M & 28.4 M & $14 \times 20$ \\
        \addlinespace
        res3-4+fpn3-5 & 40.6 & 36.2 & 78.9 M & 2.24 G & \textbf{40.7} & \textbf{36.3} & 250 M & 7.62 G & N/A \\
        res3-4+fpn1-5 & \textbf{41.2} & \textbf{36.7} & 354 M & 8.85 G & OOM & OOM & 22.4 G & 712 G & N/A \\
        \bottomrule
    \end{tabular}
\end{table*}

\subsubsection{Comparison with competing methods}
\label{sec:vs-comp}

Table \ref{tab:vs-comp} shows the comparison of absolute performance and performance improvement with competing approaches on MS-COCO 2017 object detection and instance segmentation. EA models has the highest performance and performance improvement in all settings while using the least resources. Note that EA's baseline models are significantly stronger, which make the improvements more valuable.

\begin{table*}
    \centering
    \caption{\textbf{Comparison vs. competing methods on MS-COCO 2017 object detection and instance segmentation.} For each model, the number outside the parentheses is the AP, and the number inside is the AP improvement over baseline. The table reports number of parameters and amount of computation as a percentage increase over the baseline Mask R-CNN. The team obtained these metrics by measuring the official open-source implementations of \cite{latentgnn,cgnl}. The table does not report results for CGNL with ResNet-101 and ResNeXt-101 since \cite{cgnl} did not report such results. The Table omits parameters and computation for instance segmentation since all methods modified the backbone, which the bounding box and the instance mask branches share. Therefore, the table reports the total parameter and computation change only in the rows for object detection to avoid repetition}
    \label{tab:vs-comp}
    \begin{tabular}{llrrrrr}
        \toprule
        AP type & Method & ResNet-50 & ResNet-101 & ResNeXt-101  & Parameters & Computation \\
        \midrule
        Box & EA & (+\textbf{1.8}) \textbf{41.2} & (\textbf{+1.8}) \textbf{43.1} & (\textbf{+1.4}) \textbf{44.9} & \textbf{+2.9}\% & \textbf{+5.3}\% \\ 
         & LatentGNN \cite{latentgnn} & (+1.7) 39.5 & (+1.5) 41.0 & (+1.1) 43.2 & +11.1\% & +7.6\% \\
         & CGNL \cite{cgnl} & (+1.2) 35.7 & - & - & +21.7\% & +5.7\% \\
        \addlinespace
        Mask & EA & (\textbf{+1.6}) \textbf{36.7} & (\textbf{+1.3}) \textbf{37.9} & (\textbf{+1.0}) \textbf{39.5} & - & - \\
         & LatentGNN \cite{latentgnn} & (+1.2) 35.4 & (\textbf{+1.3}) 37.2 & (\textbf{+1.0}) 38.8 & - & - \\
         & CGNL \cite{cgnl} & (+0.8) 31.2 & - & - & - & - \\
        \bottomrule
    \end{tabular}
\end{table*}

\subsection{Ablation studies}
\label{sec:ablation}

\subsubsection{Attention normalization}

These experiments empirically compared the two methods Section \ref{sec:ea} specified, namely scaling and softmax normalization. Table \ref{tab:norm} reports the experimental outcomes. The results demonstrate that the effectiveness does not depend on the specific normalization method. Following \cite{nonlocal}, all other experiments used softmax normalization.

\begin{table}
    \centering
    \caption{\textbf{Experiments on attention normalization methods on MS-COCO 2017 object detection and instance segmentation.} Experiments inserted efficient attention modules at fpn1-5}
    \label{tab:norm}
    \vspace{5pt}
    \begin{tabular}{lrr}
        \toprule
        Method & Box AP & Mask AP \\ \midrule
        Scaling & \textbf{40.2} & 35.9 \\
        Softmax & \textbf{40.2} & \textbf{36.0} \\
        \bottomrule
    \end{tabular}
\end{table}

\subsubsection{Dimensionality of the keys}
\label{sec:channels}

These experiments tested the impact of the dimensionality of the keys on the effect of efficient attention. As in Table \ref{tab:channels}, decreasing the dimensionality of the keys from 128 to 32 caused minimal accuracy change. This result reinforces the hypothesis in Section \ref{sec:intro} that most attention maps are expressible as linear combinations of a limited set of template attention maps. Therefore, researchers can reduce the dimensionality of the keys and queries in efficient attention modules to further save resources.

\begin{table}
    \centering
    \caption{\textbf{Experiments on the dimensionality of the keys on MS-COCO 2017 object detection and instance segmentation.} Experiments inserted efficient attention modules at res3-4+fpn3-5}
    \label{tab:channels}
    \vspace{5pt}
    \begin{tabular}{rrr}
        \toprule
        $d_k$ & Box AP & Mask AP \\ \midrule
        32 & 40.4 & 36.1 \\
        64 & \textbf{40.6} & \textbf{36.2} \\
        128 & 40.3 & 36.1 \\
        \bottomrule
    \end{tabular}
\end{table}

\section{Experiments on other tasks}

\subsection{Stereo depth estimation}

The experiments on efficient attention for stereo depth estimation used the Scene Flow dataset, a large-scale synthesized dataset with 39824 stereo frame pairs. The baseline is PSMNet \cite{psmnet}, a clean model with near state-of-the-art performance. The experiments empirically determined the optimal hyperparamters, which significantly outperform the setting in \cite{psmnet} (batch size is 24, learning rate is $2 \times 10^{-3}$, training length is 100 epochs, and the rest is the same as in \cite{psmnet}), as in Table \ref{tab:vs-ca-stereo}. On top of the strong baseline, inserting an efficient attention module after the last 3D hourglass leads to further improvement and achieves a new state-of-the-art. In comparison, inserting a non-local module at the same place would require an astronomical 9.68 TB of memory, prohibiting any attempt to verify its effectiveness. Table \ref{tab:vs-sota-stereo} compares EA-PSMNet with other state-of-the-art approaches and shows that it substantially outperforms all competing methods.

\begin{table*}[h!]
    \centering
    \caption{\textbf{Experiments on THUMOS14 temporal action localization.} \textit{mAP@x} stands for mean average precision at IoU threshold $x$. \textit{EA R-C3D} is this paper's model. Both models used ResNet-50 as the backbone}
    \label{tab:temp-act}
    \vspace{5pt}
    \begin{tabular}{lrrrrrrr}
        \toprule
        Model & mAP@0.1 & mAP@0.2 & mAP@0.3 & mAP@0.4 & mAP@0.5 & mAP@0.6 & mAP@0.7 \\ \midrule
        R-C3D & 54.2 & 54.1 & 50.0 & 45.6 & 37.3 & 29.2 & 18.5 \\
        \textbf{EA R-C3D} & \textbf{60.3} & \textbf{59.8} & \textbf{56.8} & \textbf{51.3} & \textbf{43.4} & \textbf{33.2} & \textbf{21.8} \\
        \bottomrule
    \end{tabular}
\end{table*}

\begin{table}
    \centering
    \caption{\textbf{Experiments on Scene Flow stereo depth estimation.} \textit{EPE} stands for end-point error and is lower the better. \textit{EA-PSMNet} is this paper's model. \textit{OOM} indicates out of memory. \textit{Memory} only counts the attention module}
    \label{tab:vs-ca-stereo}
    \vspace{5pt}
    \begin{tabular}{lrr}
        \toprule
        Model & EPE & Memory\\ \midrule
        PSMNet (original) & 1.09 & 0 \\
        PSMNet (baseline) & 0.51 & 0\\
        \textbf{EA-PSMNet} & \textbf{0.48} & 796 MB \\
        Nonlocal-PSMNet & OOM & 9.68 TB \\
        \bottomrule
    \end{tabular}
\end{table}

\begin{table}
    \centering
    \caption{\textbf{Comparison with the state-of-the-art on Scene Flow stereo depth estimation.} \textit{EPE} stands for end-point error and is lower the better. \textit{EA-PSMNet} is this paper's model}
    \label{tab:vs-sota-stereo}
    \vspace{5pt}
    \begin{tabular}{lr}
        \toprule
        Model & EPE \\ \midrule
        iResNet-i2 \cite{crl} & 1.40 \\
        EdgeStereo \cite{edgestereo} & 1.12 \\
        PSMNet \cite{psmnet} & 1.09 \\
        CSPN \cite{cspn} & 0.78 \\
        LEAStereo \cite{leastereo} & 0.78 \\
        \textbf{EA-PSMNet} & \textbf{0.48} \\
        \bottomrule
    \end{tabular}
\end{table}

\subsection{Temporal action localization}

This section presents experiments for temporal action localization on the THUMOS14 \cite{thumos14} dataset. The baseline is R-C3D \cite{rc3d}. The experiment added two efficient attention modules after \textit{res3} and \textit{res4} in the ResNet-50 backbone.  Table \ref{tab:temp-act} presents the results. At the table shows, efficient attention substantially improved the performance for this task.

\section{Visualization}
\label{sec:vis}

\begin{figure}[t]
    \centering
    \includegraphics[width=\linewidth]{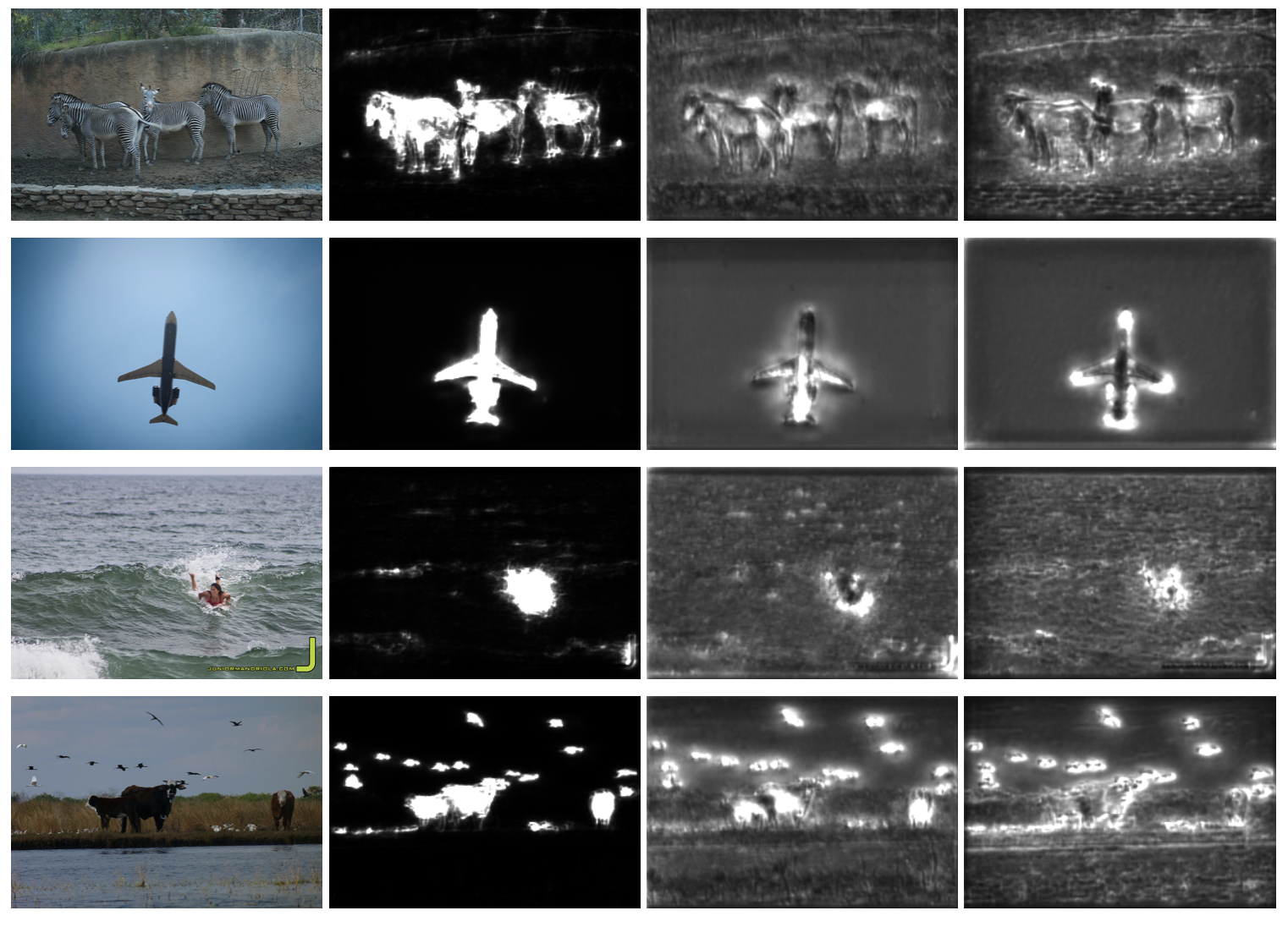}
    \caption{\textbf{Visualization of global attention maps.} The left-most column displays 4 images from MS-COCO 2017. The other three columns show three of the corresponding global attention maps from the efficient attention module at FPN level 1 for each respective example.}
    \label{fig:coco}
\end{figure}

Figure \ref{fig:coco} shows visualization of the global attention maps for various examples from the efficient attention module at fpn1 in the model corresponding to the last row in Table \ref{tab:vs-ca}. The figure illustrates 3 sets of global attention maps each with a distinct, semantic focus. Column 2 tends to capture the foreground, column 3 tends to capture the core parts of objects, and column 4 tends to capture the peripheral of objects. The semantic distinctiveness of each set of global attention maps supports the analysis in Section \ref{sec:intro} that the attention maps are linear combinations of a set of template attention maps each focusing on a semantically significant area.

\section{Conclusion}
This paper has presented the efficient attention mechanism, an attention mechanism that is quadratically more memory- and computationally-efficient than the widely adopted dot-product attention mechanism. By dramatically reducing the resource usage, efficient attention enables a large number of new use cases of attention, particularly in domains with tight resource constraints or large inputs.

The experiments verified its effectiveness on four distinct tasks, object detection, instance segmentation, and stereo depth estimation. It brought significant improvement for each task. On object detection and stereo depth estimation, efficient attention-augmented models have set new states-of-the-art. Besides the tasks this paper evaluated efficient attention on, it has promising potential in other fields where attention has demonstrated effectiveness. These fields include generative adversarial image modeling \cite{sagan,biggan} and most tasks in natural language processing \cite{transformer,gpt,bert,gpt2}. Future plans include generalizing efficient attention to these fields, as well as other fields where the prohibitive costs have been preventing the application of attention.

{\small
    \bibliographystyle{ieee_fullname}
    \bibliography{paper}

\begin{thebibliography}{10}\itemsep=-1pt

\bibitem{attention}
Dzmitry Bahdanau, Kyunghyun Cho, and Yoshua Bengio.
\newblock Neural machine translation by jointly learning to align and
  translate.
\newblock In {\em ICLR}, 2015.

\bibitem{biggan}
Andrew Brock, Jeff Donahue, and Karen Simonyan.
\newblock Large scale gan training for high fidelity natural image synthesis.
\newblock {\em arXiv preprint arXiv:1809.11096}, 2018.

\bibitem{gcnet}
Yue Cao, Jiarui Xu, Stephen Lin, Fangyun Wei, and Han Hu.
\newblock Gcnet: Non-local networks meet squeeze-excitation networks and
  beyond.
\newblock In {\em ICCV}, 2019.

\bibitem{psmnet}
Jia-Ren Chang and Yong-Sheng Chen.
\newblock Pyramid stereo matching network.
\newblock In {\em CVPR}, 2018.

\bibitem{cspn}
Xinjing Cheng, Peng Wang, and Ruigang Yang.
\newblock Learning depth with convolutional spatial propagation network.
\newblock {\em arXiv preprint arXiv:1810.02695}, 2018.

\bibitem{leastereo}
Xuelian Cheng, Yiran Zhong, Mehrtash Harandi, Yuchao Dai, Xiaojun Chang, Tom
  Drummond, Hongdong Li, and Zongyuan Ge.
\newblock Hierarchical neural architecture search for deep stereo matching.
\newblock {\em arXiv preprint arXiv:2010.13501}, 2020.

\bibitem{bert}
Jacob Devlin, Ming-Wei Chang, Kenton Lee, and Kristina Toutanova.
\newblock Bert: Pre-training of deep bidirectional transformers for language
  understanding.
\newblock {\em arXiv preprint arXiv:1810.04805}, 2018.

\bibitem{speech-tfm}
Linhao Dong, Shuang Xu, and Bo Xu.
\newblock Speech-transformer: A no-recurrence sequence-to-sequence model for
  speech recognition.
\newblock In {\em ICASSP}, 2018.

\bibitem{can}
Ruibing Hou, Hong Chang, MA Bingpeng, Shiguang Shan, and Xilin Chen.
\newblock Cross attention network for few-shot classification.
\newblock In {\em NeurIPS}, pages 4003--4014, 2019.

\bibitem{senet}
Jie Hu, Li Shen, and Gang Sun.
\newblock Squeeze-and-excitation networks.
\newblock In {\em CVPR}, 2018.

\bibitem{silco}
Tao Hu, Pascal Mettes, Jia-Hong Huang, and Cees~GM Snoek.
\newblock Silco: Show a few images, localize the common object.
\newblock In {\em ICCV}, pages 5067--5076, 2019.

\bibitem{thumos14}
Y.-G. Jiang, J. Liu, A. Roshan~Zamir, G. Toderici, I. Laptev, M. Shah, and R.
  Sukthankar.
\newblock {THUMOS} challenge: Action recognition with a large number of
  classes.
\newblock \url{http://crcv.ucf.edu/THUMOS14/}, 2014.

\bibitem{deraining}
Guanbin Li, Xiang He, Wei Zhang, Huiyou Chang, Le Dong, and Liang Lin.
\newblock Non-locally enhanced encoder-decoder network for single image
  de-raining.
\newblock In {\em ACMMM}, 2018.

\bibitem{video-reid}
Xingyu Liao, Lingxiao He, and Zhouwang Yang.
\newblock Video-based person re-identification via 3d convolutional networks
  and non-local attention.
\newblock {\em arXiv preprint arXiv:1807.05073}, 2018.

\bibitem{fpn}
Tsung-Yi Lin, Piotr Doll{\'a}r, Ross~B Girshick, Kaiming He, Bharath Hariharan,
  and Serge~J Belongie.
\newblock Feature pyramid networks for object detection.
\newblock In {\em CVPR}, 2017.

\bibitem{nlrn}
Ding Liu, Bihan Wen, Yuchen Fan, Chen~Change Loy, and Thomas~S Huang.
\newblock Non-local recurrent network for image restoration.
\newblock {\em arXiv preprint arXiv:1806.02919}, 2018.

\bibitem{crl}
Jiahao Pang, Wenxiu Sun, Jimmy~SJ Ren, Chengxi Yang, and Qiong Yan.
\newblock Cascade residual learning: A two-stage convolutional neural network
  for stereo matching.
\newblock In {\em ICCV 2017 Workshops}, 2017.

\bibitem{deep-san}
Ngoc-Quan Pham, Thai-Son Nguyen, Jan Niehues, Markus Muller, and Alex Waibel.
\newblock Very deep self-attention networks for end-to-end speech recognition.
\newblock {\em arXiv preprint arXiv:1904.13377}, 2019.

\bibitem{gpt}
Alec Radford, Karthik Narasimhan, Tim Salimans, and Ilya Sutskever.
\newblock Improving language understanding by generative pre-training.
\newblock {\em OpenAI Blog}, 2018.

\bibitem{gpt2}
Alec Radford, Jeffrey Wu, Rewon Child, David Luan, Dario Amodei, and Ilya
  Sutskever.
\newblock Language models are unsupervised multitask learners.
\newblock {\em OpenAI Blog}, 2019.

\bibitem{edgestereo}
Xiao Song, Xu Zhao, Hanwen Hu, and Liangji Fang.
\newblock Edgestereo: A context integrated residual pyramid network for stereo
  matching.
\newblock In {\em ACCV}, 2018.

\bibitem{transformer}
Ashish Vaswani, Noam Shazeer, Niki Parmar, Jakob Uszkoreit, Llion Jones,
  Aidan~N Gomez, {\L}ukasz Kaiser, and Illia Polosukhin.
\newblock Attention is all you need.
\newblock In {\em NIPS}, 2017.

\bibitem{nonlocal}
Xiaolong Wang, Ross Girshick, Abhinav Gupta, and Kaiming He.
\newblock Non-local neural networks.
\newblock In {\em CVPR}, 2018.

\bibitem{cbam}
Sanghyun Woo, Jongchan Park, Joon-Young Lee, and In~So Kweon.
\newblock Cbam: Convolutional block attention module.
\newblock In {\em ECCV}, 2018.

\bibitem{rc3d}
Huijuan Xu, Abir Das, and Kate Saenko.
\newblock R-c3d: Region convolutional 3d network for temporal activity
  detection.
\newblock In {\em ICCV}, 2017.

\bibitem{xlnet}
Zhilin Yang, Zihang Dai, Yiming Yang, Jaime Carbonell, Ruslan Salakhutdinov,
  and Quoc~V Le.
\newblock Xlnet: Generalized autoregressive pretraining for language
  understanding.
\newblock {\em arXiv preprint arXiv:1906.08237}, 2019.

\bibitem{cgnl}
Kaiyu Yue, Ming Sun, Yuchen Yuan, Feng Zhou, Errui Ding, and Fuxin Xu.
\newblock Compact generalized non-local network.
\newblock In {\em NeurIPS}, 2018.

\bibitem{sagan}
Han Zhang, Ian Goodfellow, Dimitris Metaxas, and Augustus Odena.
\newblock Self-attention generative adversarial networks.
\newblock {\em arXiv preprint arXiv:1805.08318}, 2018.

\bibitem{latentgnn}
Songyang Zhang, Shipeng Yan, and Xuming He.
\newblock {L}atent{GNN}: Learning efficient non-local relations for visual
  recognition.
\newblock In {\em ICML}, 2019.

\end{thebibliography}
}

\appendix

\section{Architecture details for experiments on MS-COCO 2017}
\label{sec:arch}

Table \ref{tab:arch} details the architecture the experiments used on MS-COCO 2017.

\begin{table*}
    \centering    
    \caption{\textbf{Architecture details for experiments on MS-COCO 2017 object detection and instance segmentation.} This table assumes the backbone architecture is ResNet-50. For ResNet-101 and ResNeXt-101, the only difference will be the number of ResBlocks in each ResBlock group (res1-4) and/or the type of the blocks (ResNeXtBlock (32x4d) instead of ResBlock)}
    \label{tab:arch}
    \begin{tabular}{lllr}
        \toprule
        Block & Type & Input & Output size \\ \midrule
        input & Input & N/A & $896 \times 1280$ \\
        conv1 & Conv $3 \times 3$ & input & $448 \times 640$ \\
        maxpool & Maxpool $2 \times 2$ & conv1 & $224 \times 320$ \\
        \addlinespace
        res1 & ResBlock $\times$ 3 & maxpool & $224 \times 320$ \\
        res2 & ResBlock $\times$ 4 & res1 & $112 \times 160$ \\
        res3 & ResBlock $\times$ 6 & res2 & $56 \times 80$ \\
        res4 & ResBlock $\times$ 3 & res3 & $28 \times 40$ \\
        \addlinespace
        fpn5 & conv $3 \times 3$ & res4 & $14 \times 20$ \\
        fpn4 & conv $3 \times 3$ & res4 + fpn5 (upsampled) & $28 \times 40$ \\
        fpn3 & conv $3 \times 3$ & res3 + fpn4 (upsampled) & $56 \times 80$ \\
        fpn2 & conv $3 \times 3$ & res2 + fpn3 (upsampled) & $112 \times 160$ \\
        fpn1 & conv $3 \times 3$ & res1 + fpn2 (upsampled) & $224 \times 320$ \\
        \addlinespace
        rpn & RPN & fpn1-4 & N/A \\
        roi & RoI Align & fpn1-4 & N/A \\
        \bottomrule
    \end{tabular}
\end{table*}

\section{Fine-grain metrics for experiments on MS-COCO 2017}
\label{tab:breakdown}

Table \ref{tab:breakdown-box} presents fine-grain object detection metrics on MS-COCO 2017. Table \ref{tab:breakdown-mask} presents find-grain instance segmentation metrics on MS-COCO 2017.

\begin{table*}
    \centering    
    \caption{\textbf{Fine-grain metrics for experiments on MS-COCO 2017 object detection.} \textit{+n NL} means adding \textit{n} non-local \cite{nonlocal} blocks to the backbone and FPN. \textit{+n EA} means adding \textit{n} EA modules to the backbone and FPN. \textit{OOM} indicates out-of-memory errors}
    \label{tab:breakdown-box}
    \begin{tabular}{lrrrrrrrrr}
        \toprule
        Backbone & AP & AP-50 & AP-75 & AP-small & AP-medium & AP-large \\
        \midrule
        ResNet-50 & 39.4 & 60.6 & 42.8 & 24.7 & 43.0 & 50.9 \\
        \addlinespace
        +1 NL & 40.3 & 61.9 & 43.6 & 24.3 & 43.8 & 52.2 \\
        +1 EA & 40.2 & 61.9 & 43.6 & 24.9 & 44.0 & 51.5 \\
        \addlinespace
        +5 NL & 40.7 & 62.1 & 44.2 & 25.3 & 44.5 & 52.0 \\
        +5 EA & 40.6 & 62.8 & 44.2 & 25.0 & 44.6 & 52.3 \\
        \addlinespace
        +7 NL & OOM & OOM & OOM & OOM & OOM & OOM \\
        +7 EA & 41.2 & 62.7 & 44.8 & 25.8 & 44.9 & 52.5 \\
        \midrule
        ResNeXt-101 & 43.5 & 65.4 & 47.5 & 27.0 & 47.9 & 55.3 \\
        +7 EA & 44.9 & 66.8 & 48.7 & 27.1 & 49.1 & 57.6 \\
        \bottomrule
    \end{tabular}
\end{table*}

\begin{table*}
    \centering    
    \caption{\textbf{Fine-grain metrics for experiments on MS-COCO 2017 instance segmentation.} \textit{+n NL} means adding \textit{n} non-local \cite{nonlocal} blocks to the backbone and FPN. \textit{+n EA} means adding \textit{n} EA modules to the backbone and FPN. \textit{OOM} indicates out-of-memory errors}
    \label{tab:breakdown-mask}
    \begin{tabular}{lrrrrrr}
        \toprule
        Backbone & AP & AP-50 & AP-75 & AP-small & AP-medium & AP-large \\
        \midrule
        ResNet-50 & 35.1 & 57.0 & 37.2 & 19.9 & 38.1 & 47.7 \\
        \addlinespace
        +1 NL & 35.9 & 58.2 & 38.1 & 20.9 & 39.2 & 47.9 \\
        +1 EA & 36.0 & 58.0 & 38.3 & 20.3 & 39.0 & 47.9 \\
        \addlinespace
        +5 NL & 36.3 & 59.4 & 38.5 & 19.7 & 39.8 & 50.4\\
        +5 EA & 36.2 & 59.2 & 38.2 & 19.9 & 39.9 & 49.4 \\
        \addlinespace
        +7 NL & OOM & OOM & OOM & OOM & OOM & OOM \\
        +7 EA & 36.7 & 59.1 & 39.2 & 21.6 & 40.2 & 48.6 \\
        \midrule
        ResNeXt-101 & 38.5 & 61.8 & 41.0 & 21.7 & 42.5 & 51.7 \\
        +7 EA & 39.3 & 63.0 & 42.0 & 22.6 & 43.0 & 52.4 \\
        \bottomrule
    \end{tabular}
\end{table*}

\end{document}